\documentclass[10pt,twocolumn,letterpaper]{article}

\usepackage{cvpr}
\usepackage{times}
\usepackage{epsfig}
\usepackage{graphicx}
\usepackage{amsmath}
\usepackage{amssymb}
\usepackage{multirow}
\usepackage[pagebackref=true,breaklinks=true,letterpaper=true,colorlinks,bookmarks=false]{hyperref}

\cvprfinalcopy % *** Uncomment this line for the final submission

\def\cvprPaperID{1173} % *** Enter the CVPR Paper ID here

\ifcvprfinal\fi
\begin{document}

%%%%%%%%% TITLE
%\title{Quantize Fully Convolutional Networks for Edge Biomedical Image Segmentation with Improved Suggestive Annotation}
%\title{Enhanced Suggestive Annotation for Quantization of Fully Convolutional Networks towards Edge Biomedical Image Segmentation }
\title{Quantization of Fully Convolutional Networks for Accurate Biomedical\\ Image Segmentation }

\author{
Xiaowei Xu$^{1,2}$, Qing Lu$^1$, Yu Hu$^2$, Lin Yang$^1$, Sharon Hu$^1$, Danny Chen$^1$, Yiyu Shi$^1$\\
1 Univerity of Notre Dame\\
2 Huazhong University of Science and Technology\\
%For Blind Review\\
%Institution1\\
%Institution1 address\\
%{\tt\small firstauthor@i1.org}
% For a paper whose authors are all at the same institution,
% omit the following lines up until the closing ``}''.
% Additional authors and addresses can be added with ``\and'',
% just like the second author.
% To save space, use either the email address or home page, not both
%\and
%Second Author\\
%Institution2\\
%First line of institution2 address\\
%{\tt\small secondauthor@i2.org}
}

\maketitle
%\thispagestyle{empty}

%%%%%%%%% ABSTRACT
\begin{abstract}
%With the need for increased care and welfare of the rapidly aging population, mobile telemedicine is becoming popular for providing remote health care to increase the quality of life.
%In mobile telemedicine, biomedical image segmentation plays a critical role for medical diagnosis and treatment, which is best done locally or at the "edge" considering the large volume of high-resolution images and the real-time constraints that are commonly presented.
%Usually the so-called edge biomedical image segmentation is performed on embedded devices typically bounded by limited resource and energy.
%However, the current trend for promising biomedical image segmentation is based on fully convolutional networks (FCN) which is characterized with high memory (tens to hundreds of million bytes) and high computation requirement (millions to billions of operations).
%Obviously, the demand of computation and energy is far beyond that current embedded devices can provide.

With pervasive applications of medical imaging in health-care, biomedical image segmentation plays a central role in quantitative analysis, clinical diagnosis, and medical intervention.
%With the prevalence usage of medical imaging in health-care, biomedical image segmentation has played a significant role in quantitative reasoning, clinical analysis and medical intervention.
Since manual annotation suffers limited reproducibility, arduous efforts, and excessive time, automatic segmentation is desired to process increasingly larger scale histopathological data.
%Due to the limited reproducibility and considerable time and efforts of manual annotation, automatic segmentation is preferred to process the increasingly large-scale histopathological data.
%Recently, deep neural networks (DNNs), particularly fully convolutional networks (FCN) have been widely studied for biomedical image segmentation with promising performance.
Recently, deep neural networks (DNNs), particularly fully convolutional networks (FCNs), have been widely applied to biomedical image segmentation, attaining much improved performance.
%Quantization of DNNs has been emerging, which aims to represent weights with fewer bits resulting with a great reduction on memory and computation requirement.
At the same time, quantization of DNNs has become an active research topic, which aims to represent weights with less memory (precision) to considerably reduce memory and computation requirements of DNNs while maintaining acceptable accuracy.
In this paper, we apply quantization techniques to FCNs for accurate biomedical image segmentation.
%Recently, quantization of deep neural networks (DNN) has been widely adopted, which represent weights with less bits resulting with a great reduction on memory and computation requirement.
%In this paper, we approach a step further, and we quantize fully convolutional networks for edge biomedical image segmentation with enhanced suggestive annotation.
Unlike existing literature on quantization which primarily targets memory and computation complexity reduction, we apply quantization as a method to reduce overfitting in FCNs for better accuracy.
Specifically, we focus on a state-of-the-art segmentation framework, suggestive annotation \cite{yang2017suggestive}, which judiciously extracts representative annotation samples from the original training dataset, obtaining an effective small-sized balanced training dataset.
We develop two new quantization processes for this framework: (1) suggestive annotation with quantization for highly representative training samples, and (2) network training with quantization for high accuracy.
%and add two quantization processes to it: one to suggestive annotation for highly representative training samples, and the other to network training for high accuracy.
Extensive experiments on the MICCAI Gland dataset show that both quantization processes can improve the segmentation performance, and our proposed method exceeds the current state-of-the-art performance by up to 1\%.
In addition, our method has a reduction of up to 6.4x on memory usage.
%state-of-the-art performance can be achieved by combining the two.
\end{abstract}

\vspace{-10pt}
%%%%%%%%% BODY TEXT
\section{Introduction}

With pervasive applications of medical imaging in health-care, biomedical image segmentation has always been one of the most important tasks in biomedical imaging research.
Biomedical image segmentation extracts different tissues, organs, pathologies, and biological structures, to support medical diagnosis, surgical planning and treatments.
In common practice, segmentation is performed manually by pathologists, which is time-consuming and tedious.
However, the ever-increasing quantity and variety of medical images make manual segmentation impracticable in terms of cost and reproducibility.
Therefore, automatic biomedical image segmentation is highly desirable.
But, this task is very challenging,
because of high variability in medical images due to complex variations in biomedical objects and structures and because of low contrast, noise, and other imaging artifacts caused by various medical imaging modalities and techniques.

%With the pervasive usage of medical images in health-care, biomedical image segmentation has become one of the most important tasks in biomedical imaging.
%With biomedical image segmentation, different tissue classes, organs, pathologies, or other biologically relevant structures can be extracted, which can be used to support medical diagnosis, surgical planning and treatments.
%Traditionally, this process is performed manually by pathologists, which is time-consuming and tedious.
%However, the ever-increasing quantity and variety of medical images makes manual segmentation unpracticable considering cost and reproducibility.
%Therefore, automatic biomedical image segmentation is preferred.
%Nevertheless, this task is challenging for several reasons.
%First, there exists high variability in medical images due to the complex modes of variation of human anatomy.
%Second, biomedical image segmentation is made difficult by low contrast, noise, and other imaging ambiguities due to the drawback of medical imaging techniques.

In the past years, substantial progress has been made on biomedical image segmentation with pixel based methods \cite{doyle2006boosting,nguyen2012structure,tabesh2007multifeature,sirinukunwattana2015novel} and structure based methods \cite{altunbay2010color,gunduz2010automatic,fu2014novel,sirinukunwattana2015stochastic}.
These methods achieve promising results on nonmalignant objects using hand-crafted features and prior knowledge of structures.
However, they suffer considerable degradation when applied to malignant objects with serious deformation.
Recently, deep neural networks (DNNs), particularly fully convolutional networks (FCNs), have been highly effective for biomedical image segmentation, which require little hand-crafted features or prior knowledge.
Ronneberger et al. \cite{ronneberger2015u} proposed U-Net, a U-shaped deep convolutional network that adds a symmetric expanding path to enable precise localization.
With strong use of data augmentation, this segmentation model achieves significant improvement over previous methods.
The DCAN model by Chen et al. \cite{chen2016deep,chen2016dcan} added a unified multi-task object to the U-Net learning framework, which won the 2015 MICCAI Gland Segmentation Challenge \cite{sirinukunwattana2017gland}.
Based on DCAN, Yang et al. \cite{yang2017suggestive} proposed suggestive annotation which extracts representative samples as a training dataset, by adopting active learning into their network design.
%Currently, suggestive annotation achieves state-of-the-art performance for biomedical image segmentation.
With the refined training samples and optimized structure for DNNs, suggestive annotation achieves state-of-the-art performance on the MICCAI Gland Segmentation dataset \cite{sirinukunwattana2017gland}.

%cnn methods (unet, dcan, suggestive annotation)

%Recent advances in deep neural networks (DNNs), or more specifically FCN \cite{ronneberger2015u}\cite{chen2016deep}\cite{chen2016dcan}\cite{yang2017suggestive} have achieved promising results on many biomedical image segmentation benchmarks \cite{arganda2015crowdsourcing}\cite{sirinukunwattana2017gland}.
%FCN has become a main choice for biomedical image segmentation for its accuracy and generality.
%However, FCN suffers from high memory and high computation requirement in which there are millions of parameters and billions of operations are required to process each image.
%Thus, there exists a big gap between the required performance and the limited resource that a "edge" device or an embedded device can provide.

At the same time, DNN quantization has become an active research topic, which aims to represent DNN weights with less memory (precision) while maintaining acceptable accuracy with efficient memory and computation costs.
It has been observed in the literature, however, that sometimes quantization can improve accuracy which can be credited to the reduction of overfitting.
%low-bit representation while maintaining an acceptable accuracy for efficient memory access and computation.
%Precision reduction.
Dynamic fixed point are adopted in \cite{han2015deep}\cite{hubara2016quantized}, which achieves 4x less memory operation cost with only 0.4-0.6\% Top-5 accuracy loss for ImageNet classification \cite{deng2009imagenet}.
Ternary weight network \cite{li2016ternary} and binaryConnect \cite{courbariaux2015binaryconnect} have further reduced the bit-width of weights to 2 bits or even 1 bit with a relatively larger accuracy loss.
Recently, their enhanced version, trained ternary training \cite{zhu2016trained} and binary weight network \cite{rastegari2016xnor} have reduced the accuracy loss to only 0.6-0.8\%.
There also exists some works using non-linear quantization to represent the parameter distribution for better accuracy \cite{han2015deep}\cite{zhou2017incremental}.
Unlike the above works, some studies aims to quantize not only the weights but also the activations.
Quantized neural networks \cite{hubara2016quantized}, binarized neural networks \cite{courbariauxbinarynet}, and XNOR-net \cite{rastegari2016xnor} reduced the weights to only 1 bit and the activations to 1-2 bits resulting in a large reduction on memory and computation cost yet with significant accuracy loss.
Particularly, all the computation in XNOR-net are very simple XNOR operation.
In some of the above works, we notice that quantization can sometimes improve the performance \cite{han2015deep}\cite{zhou2017incremental}, which can be credited to the reduction of overfitting.

In this paper, we adopt quantization as a method to reduce overfitting to FCNs for accurate biomedical image segmentation.
Particularly, we focus on a recent effective biomedical image segmentation framework, suggestive annotation \cite{yang2017suggestive}.
%Particularly we base our work on a recent work \cite{yang2017suggestive}, and it has two steps: suggestive annotation with FCNs and FCN training.
We develop two new quantization processes to incorporate into this state-of-the-art framework: (1) suggestive annotation with quantization for highly representative training samples, and (2) network training with quantization for high accuracy.
%We add two quantization processes to the two steps, respectively: one to suggestive annotation for high-representative training samples, and another to network training for high accuracy.
%Suggestive annotation is a critical step which extracts representative samples from the raw training samples.
%In this paper, we first apply general quantization to FCNs to compress model.
%Note that unlike general DNNs, FCNs have no full connection layers.
%Second, we go a step further and add a quantization processing to the current suggestive annotation, which yields a better samples for training.
%Results with a variety of quantization methods show that our approach can achieves much better accuracy than general quantized FCNs.
Extensive experiments are presented on the widely-used MICCIA Gland dataset, and the results show that our proposed method exceeds the current state-of-the-art performance by up to 1\%.
In addition, our method has a reduction of up to 6.4x on memory usage.
%In addition, our method have a 4.6x-6.4x reduction on memory usage.

%-------------------------------------------------------------------------
\section{Related Work}
In this section, we briefly review suggestive annotation \cite{yang2017suggestive}, on which our proposed method is based.
Several representative quantization methods are discussed in detail, which will be adopted in our experiments.
%Other quantization methods are also discussed in the last subsection.
The readers are also referred to \cite{han2015deep,courbariaux2015binaryconnect,rastegari2016xnor} for other quantization methods.

\begin{figure}
\begin{center}
\centerline{\includegraphics[width=1\columnwidth]{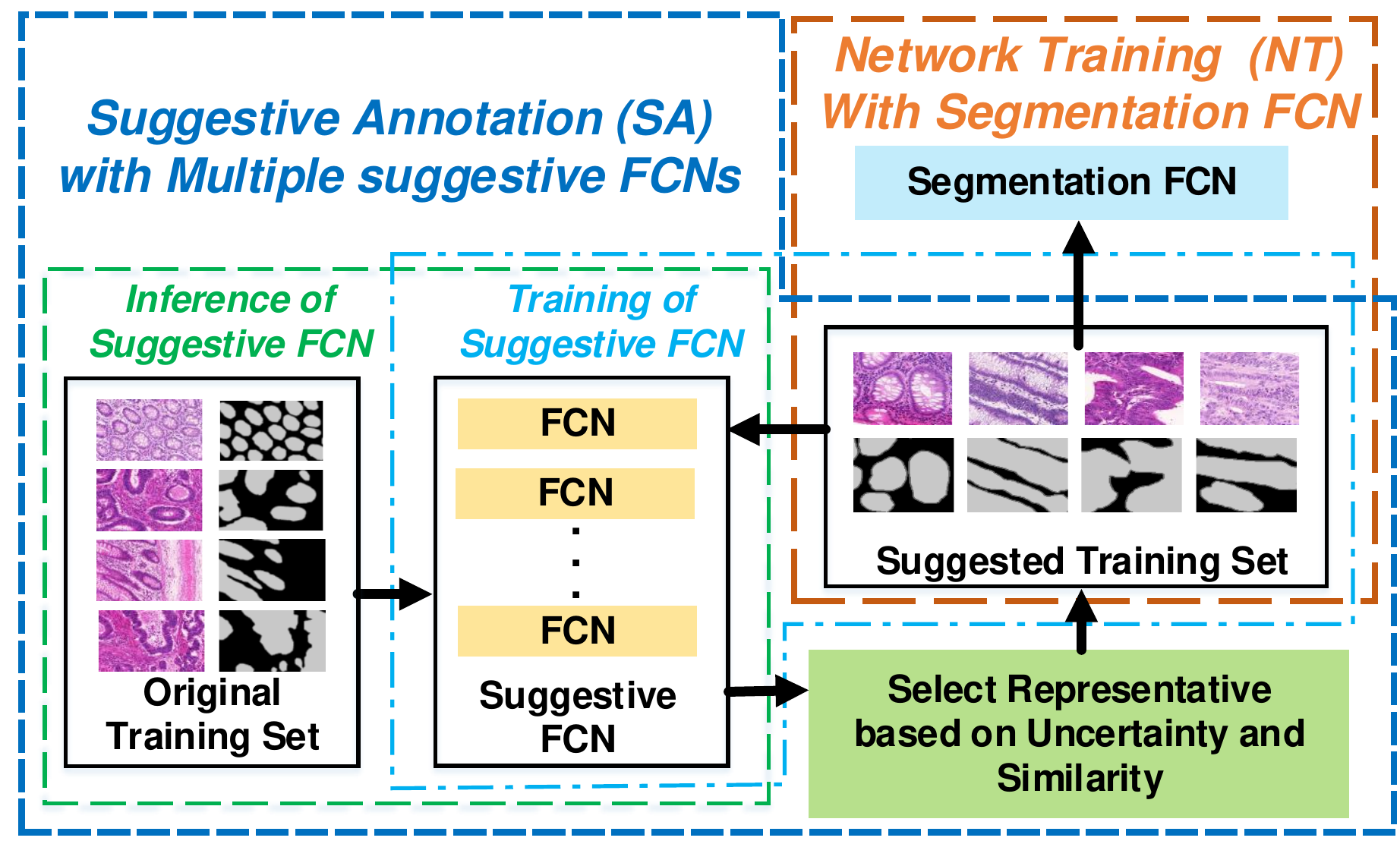}}
\end{center}
\vspace{-20pt}
   \caption{Illustration of the suggestive annotation framework \cite{yang2017suggestive}. With suggestive annotation, better training samples (suggestive training set) can be extracted from the original training set for further training with better performance.}
\label{fig:architectureOrigin}
\end{figure}

\begin{figure}%[ht]
\begin{center}
\centerline{\includegraphics[width=0.7\columnwidth]{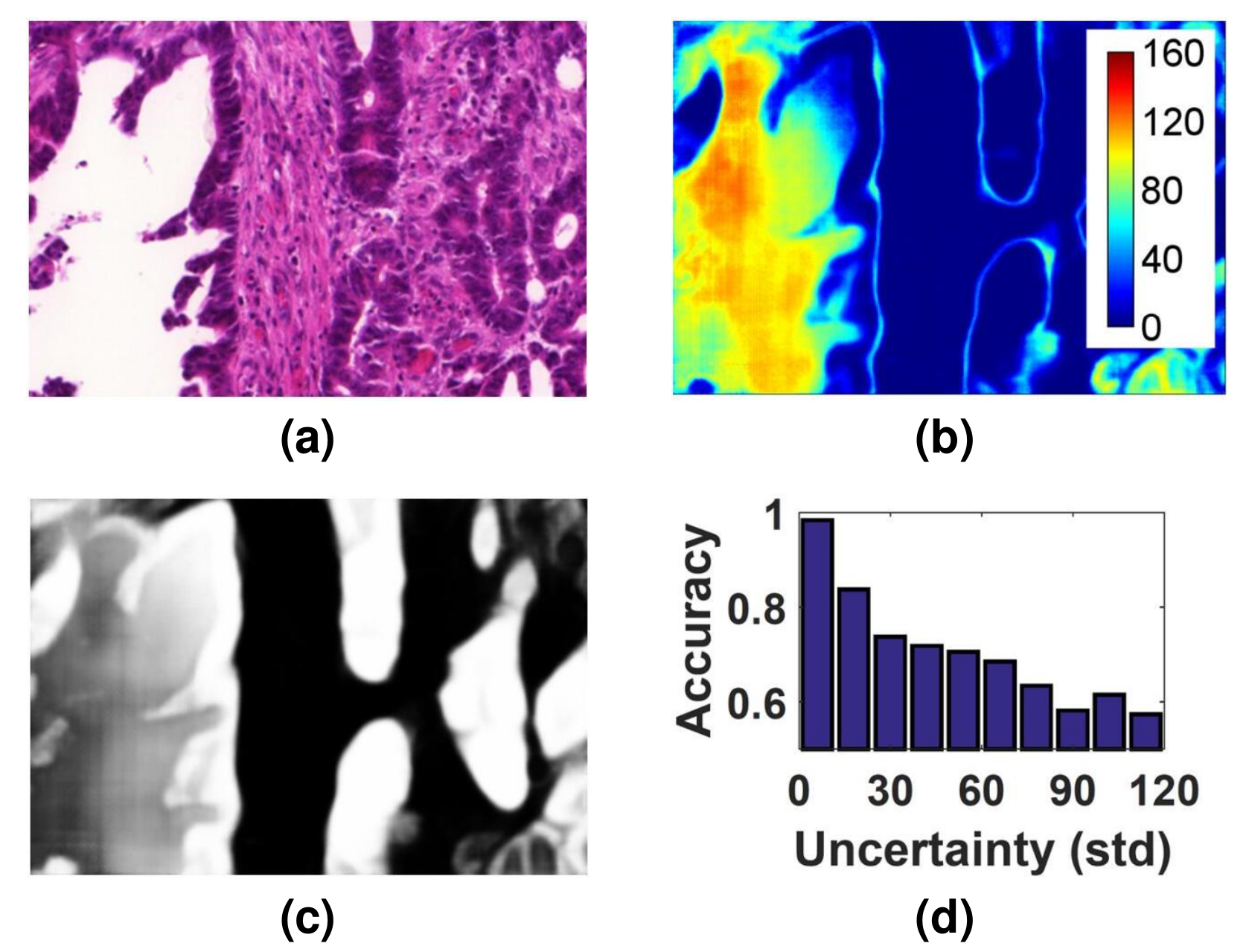}}
\end{center}
\vspace{-20pt}
   \caption{(a) An original image; (b) the probability map produced by multiple FCNs in suggestive annotation for (a);
(c) uncertainty estimation of the results by the multiple FCNs; (d) relation between uncertainty estimation
and pixel accuracy on the testing data. Obviously there is a strong correlation between the test accuracy and uncertainty (reprinted from \cite{yang2017suggestive}).}
\label{fig:correlation}
\vspace{-10pt}
\end{figure}

\subsection{Suggestive Annotation for Biomedical Image Segmentation}
We based our proposed framework on suggestive annotation \cite{yang2017suggestive}, which achieves state-of-the-art performance on the Gland dataset.
The key idea of the work is that better performance can be achieved with representative training samples instead of original training samples.
As shown in Figure \ref{fig:architectureOrigin}, the suggestive annotation framework \cite{yang2017suggestive} has two steps: suggestive annotation and network training.
The first step extracts typical samples from the original training set with multiple suggestive FCNs, and the second step trains segmentation FCNs with the extracted samples.
In the first step, multiple suggestive FCNs are trained in parallel.
During the inference stage, multiple suggestive FCNs produce multiple predictions for the same input from the original training set, which can be used to calculate the representativeness of the samples.

Note that each FCN generates two outputs: contour of the objects and the segmented object.
The suggestive FCNs and segmentation FCNs have the same network structure which is based on DCAN \cite{chen2016dcan} and active learning.
%and please refer to work \cite{yang2017suggestive} for more details of FCN implementation.

Two metrics are involved with representativeness: uncertainty and similarity.
A representative training samples should be hard to predict as they are located on the "boarder line" of the feature space, and have low similarity with each other as they can well describe the variety of the "boarder line" with limited quantity.
%Note that high uncertainty scores means the sample is more hard to predict, which therefore is more typical to describe the 'boarder line' of the space.
In suggestive annotation, the standard deviation of the multiple predictions from multiple suggestive FCNs are regarded as the uncertainty score.
The averaged outputs of last convolutional layers of multiple suggestive FCNs are regarded as a domain-specific image descriptor, which can be used to evaluate the similarity of images with cosine similarity.
%With uncertainty and similarity, the first step can extract difficult-to-segment (uncertainty) and representative with common features (similarity) samples.

Selecting representative training samples with uncertainty and similarity is an NP-hard problem \cite{yang2017suggestive}.
A simple heuristic method is adopted: extract $K$ ($k<K$) samples with the highest uncertainty scores first, and then select the final $k$ samples based on their similarity with each other.
The reason to put uncertainty in the first step is that uncertainty is more important than similarity \cite{yang2017suggestive}.
As shown in Figure \ref{fig:correlation}, the test accuracy is highly correlated with the uncertainty scores.
%Thus, in the enhanced suggestive annotation, a heuristic method is used to select the representative samples, and samples with high uncertainty are extracted first.

\subsection{Quantization Techniques for DNNs}
\subsubsection{Incremental Quantization (INQ)}

Incremental quantization \cite{zhou2017incremental} quantizes weights to powers of two in an iterative manner.
In each iteration, a subset of weights is selected and quantized, and a fine-tuning process is then presented while the quantized weights are locked during both feed-forward and feed-back prorogation.
The above process iterates until all weights are quantized.
The quantization calculation is shown in Eq. (\ref{formula:quantization1}), where $w^q$ and $w$ are quantized and original weights, respectively, and $u$ and $l$ are the upper and lower bounds of the quantized set, respectively.
Note that how to choose the weights during each iteration is dependant on the magnitude of the weight.
With incremental quantization, the weights can be represented with only 3-5 bits with almost no accuracy loss, and the multiplication can be simplified to shift operation.

\vspace{-15pt}
 {%\footnotesize
%\vspace{-3pt}
%\begin{small}
\begin{equation}\label{formula:quantization1}
w^q=\begin{cases}
sign(w)\times2^p & \text{ if } 3\times2^{p-2}\leq |w|< 3\times2^{p-1}; \\
    &                    \;\;\;\;\;\;\;\;\;\;\;\;l\leq p\leq u;\\
sign(w)\times2^m & \text{ if } |w|\geq 2^u; \\
0   & \text{ if } |w| < 2^{-l-1}.\\
\end{cases}
%uq(i)=
%\left\{\begin{matrix}
%2^i,\;\;if\;3\times2^{i-2}\leq abs(uq(i))< 3\times2^{i-1}; \\
%2^m,\;\;if\;abs(uq(i))\geq 2^m;\;\;\;\;\;\;\;\;\;\;\;\;\;\;\;\;\;\;\;\;\;\;\;\;\;\;\;\;\\
%0, \;\;if\;abs(uq(i))<2^{-m}.\;\;\;\;\;\;\;\;\;\;\;\;\;\;\;\;\;\;\;\;\;\;\;\;\;
%\end{matrix}\right.
\end{equation}
%\end{small}
%\vspace{-3pt}
}

\vspace{-20pt}
\subsubsection{DoReFa-Net}
DoReFa-Net \cite{zhou2016dorefa} trains DNNs with low bitwidth weights and activations represented using low bitwidth parameter gradients, and it enables training acceleration of low bitwidth neural network on general hardware.
In the quantization process, weights and activations can be deterministically quantized, while gradients need to be stochastically quantized.
DoReFa-Net adopts a simple quantization method to quantize 32 bits values to only 1 bits as shown in Eq. (\ref{formula:quantization2}), where $w_l$ and $w^q_l$ are the original and quantized weights of the $l$th layer, respectively, and $E(|w_l|)$ calculates the mean of the absolute value of weights in the $l$th layer.
{%\footnotesize
%\vspace{-3pt}
%\begin{small}
\begin{equation}\label{formula:quantization2}
w^q_l=E(|w_l|)\times sign(w_l)
\end{equation}
%\end{small}
%\vspace{-3pt}
}
Thus, DoReFa-Net can achieve a 32x compression rate atmost with comparable accuracy compared with floating-point networks, and the computation of multiplication is also simplified to addition and/or substraction.
%Note that DoReFa-Net can also achieve a comparable accuracy compared with floating-point networks.
In the feed-back propagation, weights and gradients are maintained in floating point, and quantized weights are only used in the feed-forward propagation.
%-------------------------------------------------------------------------
\subsubsection{Ternary Weight Networks}
TWN \cite{li2016ternary} trains DNNs with weights constrained to only three values $\pm\alpha_l$ and 0.
Compared with DoReFa-Net, TWN has an extra zero, which requires 2 bits to present weights while also improving the performance.
Note that TWN is also applied in a layer-wise manner, which is the same with DoReFa-Net.
For each layer, the quantization of TWN is shown in Eq. (\ref{formula:quantization3}).
\begin{equation}\label{formula:quantization3}
w_l^q=\begin{cases}
\alpha_l    & \text{ if } |w_l|>\delta_l; \\
0           & \text{ if } -\delta_l \leq |w| \leq \delta_l; \\
-\alpha_l   & \text{ if } |w_l|<-\delta_l;.\\
\end{cases}
\end{equation}
As there is no deterministic solution for $\delta_l$ and $\alpha_l$, an approximated optimal solution is presented as shown in Eq. (\ref{formula:quantization4}) and Eq. (\ref{formula:quantization5}).
Note that the feed-back propagation are the same as that for DoReFa-Net.

\begin{equation}\label{formula:quantization4}
\delta_l=0.7\times E(|w_l|) \\
\end{equation}
\begin{equation}\label{formula:quantization5}
\alpha_l = \underset{i\in \{ i |w_l(i))|\}>\delta_l }{E}(|w_l(i))|)
\end{equation}

\begin{figure*}[!htbp]
\begin{center}
\centerline{\includegraphics[width=1.8\columnwidth]{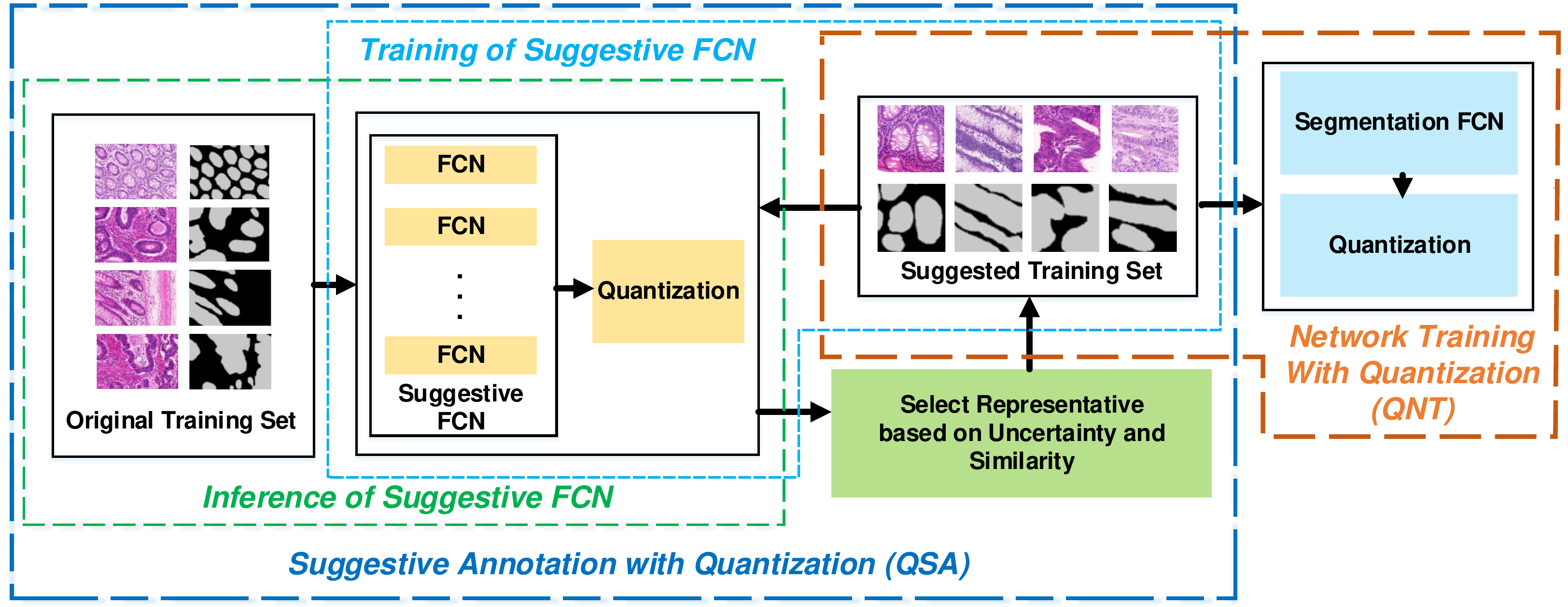}}
\end{center}
\vspace{-20pt}
   \caption{Illustration of quantization framework based on the suggestive annotation framework. In suggestive annotation with quantization, better training samples (suggestive training set) can be extracted from the original training set. In network training with quantization, better performance can be achieved by reduce overfitting. }
\label{fig:architecture}
\vspace{-10pt}
\end{figure*}
\section{Motivation}

%Usually Deep neural networks (DNNs) are over-parameterized or overfitting, and there exits a large portion of redundance in DNNs.
%To lessen the amount of overfitting, a lot of techniques are proposed to refine the training dataset and constraint the complexity of DNNs.
%Data augmentation adds more samples to the training dataset to refine it, which expands the variety of the training dataset resulting with better performance.
%In order to quantize FCNs with high compression rate and high accuracy, we motivate our work from the following two aspects: .

%This can be regarded as a technique to lessen the amount of overfitting.

Usually quantization of DNNs are used to reduce the bit length of weights in DNNs.
In fact, quantization can not only reduce memory consumption, but also can improve the performance sometimes.
For example, Han et al. \cite{han2015deep} has improved the Top-1 error by 0.01\% for ImageNet classification.
Zhou et al. \cite{zhou2017incremental} has quantized DNNs to only 4, and 5 bits for ImageNet classification, and the Top-1 and Top-5 error for the two configurations are all improved with a reduction of 0.2\%-1.47\%.
One interesting phenomenon is that the Top-5 error with quantization of 3 bits is lower than that with quantization of 4 bits.
A possible explanation is that lower bits representation is a more strict constraint to reduce overfitting.
We would like to apply the above idea to suggestive annotation \cite{yang2017suggestive} to reduce overfitting and improve performance.

Two quantization processes for the two steps in the suggestive annotation framework have different purposes.
For suggestive annotation, the purpose is to obtain representative samples, and therefore, uncertainty is more critical than accuracy.
For network training, the purpose is to increase accuracy, and several characteristics of FCNs need to be considered.
First, unlike general DNNs with multiple fully connected layers, all layers in FCNs are convolutional or deconvolutional layers, which is an extreme case of weight sharing.
Second, unlike general classification tasks with only several outputs, FCNs generates the same number of outputs as that of the inputs.
This makes quantization of FCNs much harder, which has less space for quantization compared with general DNNs.
We would like to explore suitable quantization method for FCNs in network training.

%the refinement of training dataset is achieved by extract representative samples with suggestive FCNs, which produces a more balanced training dataset.
%We believe that there still exists much redundance in suggestive FCNs, and we would like to produce more representative samples with quantization.

%As several techniques to reduce the complexity of DNNs such as weight sharing, batch normalization, dropout, and ensemble methods have been already adopted, we would like to use quantization methods to reduce overfitting.

%Suggestive annotation also adopts several techniques to reduce the complexity of DNNs such as weight sharing, batch normalization, dropout, and ensemble methods.
%Note that convolutional layer belongs to weight sharing technique.
%Another motivation is the unique characteristics compared with that of general DNNs.
%When it comes to quantization of FCNs in segmentation annotation, there are several characteristics compared with quantization of general DNNs.

\section{Method}
In this section, the proposed method is discussed in details.
As network training with quantization is relatively simple, we focus on suggestive annotation with quantization.
Additionally, uncertainty and similarity of the enhanced suggestive annotation are also analysed in details.

\subsection{Suggestive Annotation with Quantization}

As shown in Figure \ref{fig:architecture}, the proposed quantization framework has two steps: suggestive annotation with quantization and network training with quantization.
In the first step, we add a quantization module to suggestive FCNs for high uncertainty.
In the second step, quantization of segmentation FCNs are performed with the suggestive training samples for higher accuracy.
%Usually a general quantization way is to quantize the segmentation FCN only as results from FCNs with floating-point representation is with high quality.
%We question such an assumption and we also add an quantization process in the first step for better suggestive training set.
%Our intuitions are as follows.
%Unlike segmentation FCNs or general DNNs, suggestive FCNs focus on uncertainty rather than accuracy.
%At the same time,
In order to obtain high representativeness, each FCNs in suggestive FCNs should be diverse for high uncertainty with acceptable accuracy.
However, usually DNNs including FCNs are over-parameterized, and a large portion of the parameters is redundant.
Thus, multiple suggestive FCNs will have very small variance of the final prediction though with different weight initialization.
The adopted regularization techniques including weight decay and dropout scheme will further make the multiple suggestive FCNs to be almost the same.
%In fact, we could use a relatively simple FCN as suggestive FCN.
%However, considering the minor difference on behaviours between different network structures and the extra cost, we tend to use the same FCNs for suggestive FCNs with segmentation FCNs.
%Popular regularization techniques should also be valid for robust models.
%Finally, we come up with the adoption of parameter quantization, which constraint the parameter space to defense overfitting problem.
By adding quantization to suggestive annotation, the above requirement can be satisfied.
Though it may be a little offensive since most of the time it will degrade the accuracy, it is particularly appreciated by suggestive FCNs that focus on uncertainty.
Note that accuracy should be also considered and too offensive quantization methods should be avoided.
%In this paper, we adopted incremental quantization \cite{zhou2017incremental} which has very little impact on accuracy.
\begin{figure*}[!htbp]
\begin{center}
\centerline{\includegraphics[width=2\columnwidth]{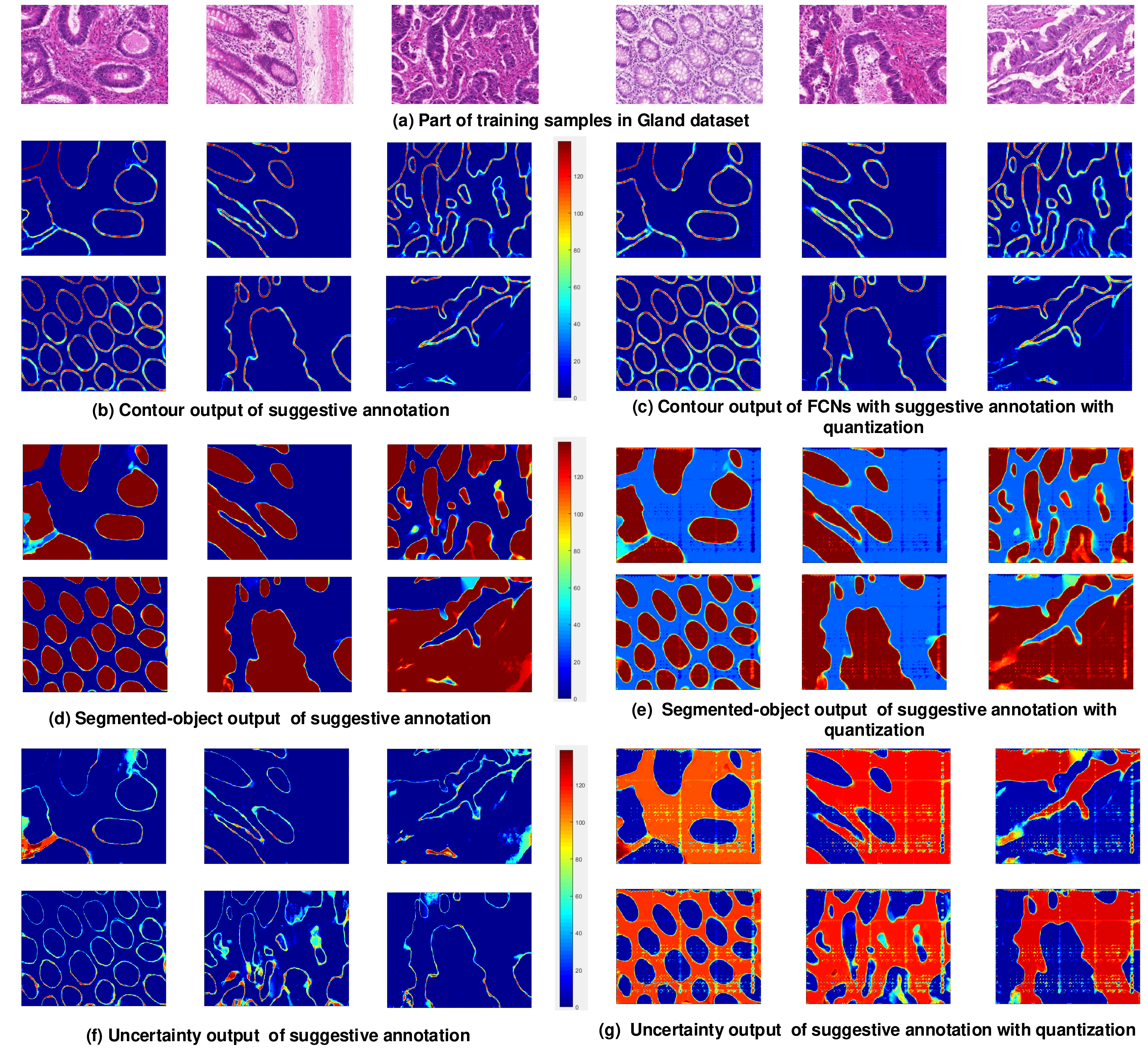}}
\end{center}
\vspace{-20pt}
   \caption{Uncertainty comparison between suggestive annotation and suggestive annotation with quantization. The accuracy of contour and segmented object and uncertainty are compared, respectively. There is almost no accuracy loss. However, suggestive annotation with quantization has higher uncertainty scores.}
\label{fig:uncertainty}
\end{figure*}

\begin{figure*}[!htbp]
\begin{center}
\centerline{\includegraphics[width=2\columnwidth]{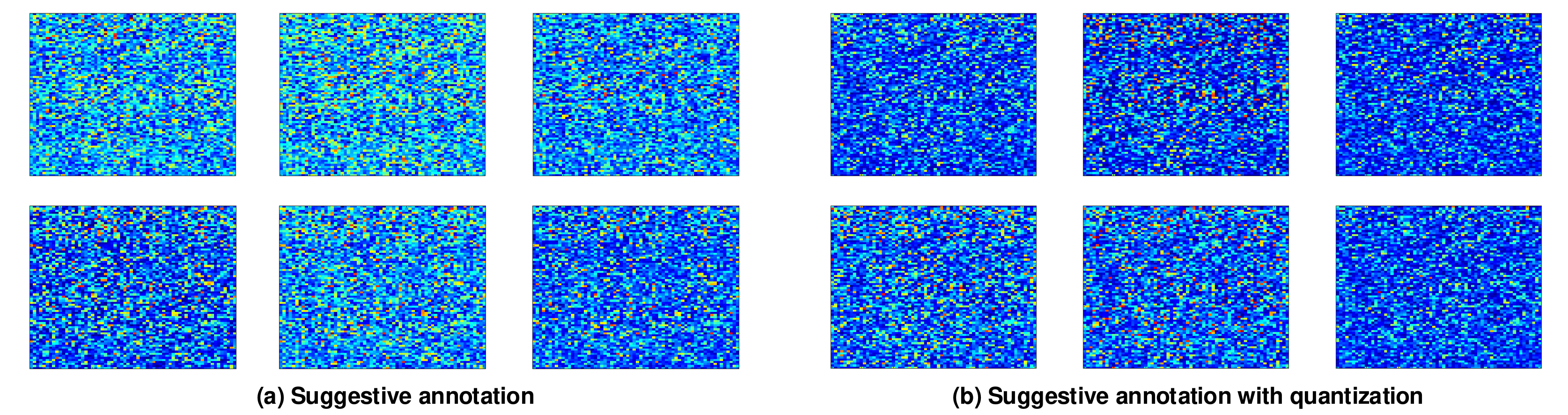}}
\end{center}
\vspace{-20pt}
   \caption{Similarity comparison between suggestive annotation and suggestive annotation with quantization.}
\label{fig:similarity}
\end{figure*}

%It should be emphasized that the above enhanced suggestive annotation is particularly useful for recent DNNs training.
%First, with enhanced suggestive annotation, more typical samples could be extracted, and higher performance can be achieved.
%Second, recent DNNs suffers from large training dataset and tedious training time.
%Enhanced suggestive annotation can reduce redundance in the training dataset and only a small dataset are used for training, which will also shorten the training time.
%The key issue here is how to adopt it to other applications.
%Biomedical image segmentation has a lot of common and repeated structures, however, other applications such as classification not.
%This will be our future work.

%Compared with general quantization ways, we go a step further.

%somekind like a defeat-over-fitting method: too overfitting, no clear uncertainty; too few parameters, no reasonable outputs;

\subsection{Impact on Uncertainty and Similarity}\label{samples}
In suggestive annotation with quantization, high uncertainty can be obtained without sacrificing much accuracy.
As shown in Figure \ref{fig:uncertainty}, accuracy including contour and segmented object and uncertainty are compared.
Note that the suggestive FCNs output both contour and segmented object for high segmentation performance.
Comparing Figure \ref{fig:uncertainty}(b) and Figure \ref{fig:uncertainty}(c), we can notice that the contour for both approaches are almost the same, and they can both obtain clear contours.
However, for segmented object in Figure \ref{fig:uncertainty}(d) and Figure \ref{fig:uncertainty}(e), suggestive annotation identifies a very clear segmented object, while the quantized version is relatively vague.
This is mainly due to the fact that suggestive annotation with quantization has a larger uncertainty of the background data, and this is verified in Figure \ref{fig:uncertainty}(f) and Figure \ref{fig:uncertainty}(g).
The uncertainty scores of suggestive annotation with quantization are much higher than that of suggestive annotation.
Therefore, suggestive training set with higher uncertainty can be obtained with quantization at the same time with little accuracy loss.

As shown in Figure \ref{fig:similarity}, the similarity (the output of the last convolutional layer) comparison between suggestive annotation and suggestive annotation with quantization is discussed.
As the dimension of the output image matrix is relatively large (64$\times$80), the details of it are not clear.
However, we can still notice that the distributions of the image of the two approaches have some differences.
For each approach, there also exists variance among the outputs.

%Note that uncertainty is more important than similarity \cite{yang2017suggestive}.
%As shown in Figure \ref{fig:correlation}, the test accuracy is highly correlated with the uncertainty.
%Thus, in the enhanced suggestive annotation, a heuristic method is used to select the representative samples, and samples with high uncertainty are extracted first.

%Uncertainty discussion (a big figure, double column),
%similarity discussion (distribution);

%error with uncertainty;

\begin{figure}%[!htbp]
\begin{center}
\centerline{\includegraphics[width=1\columnwidth]{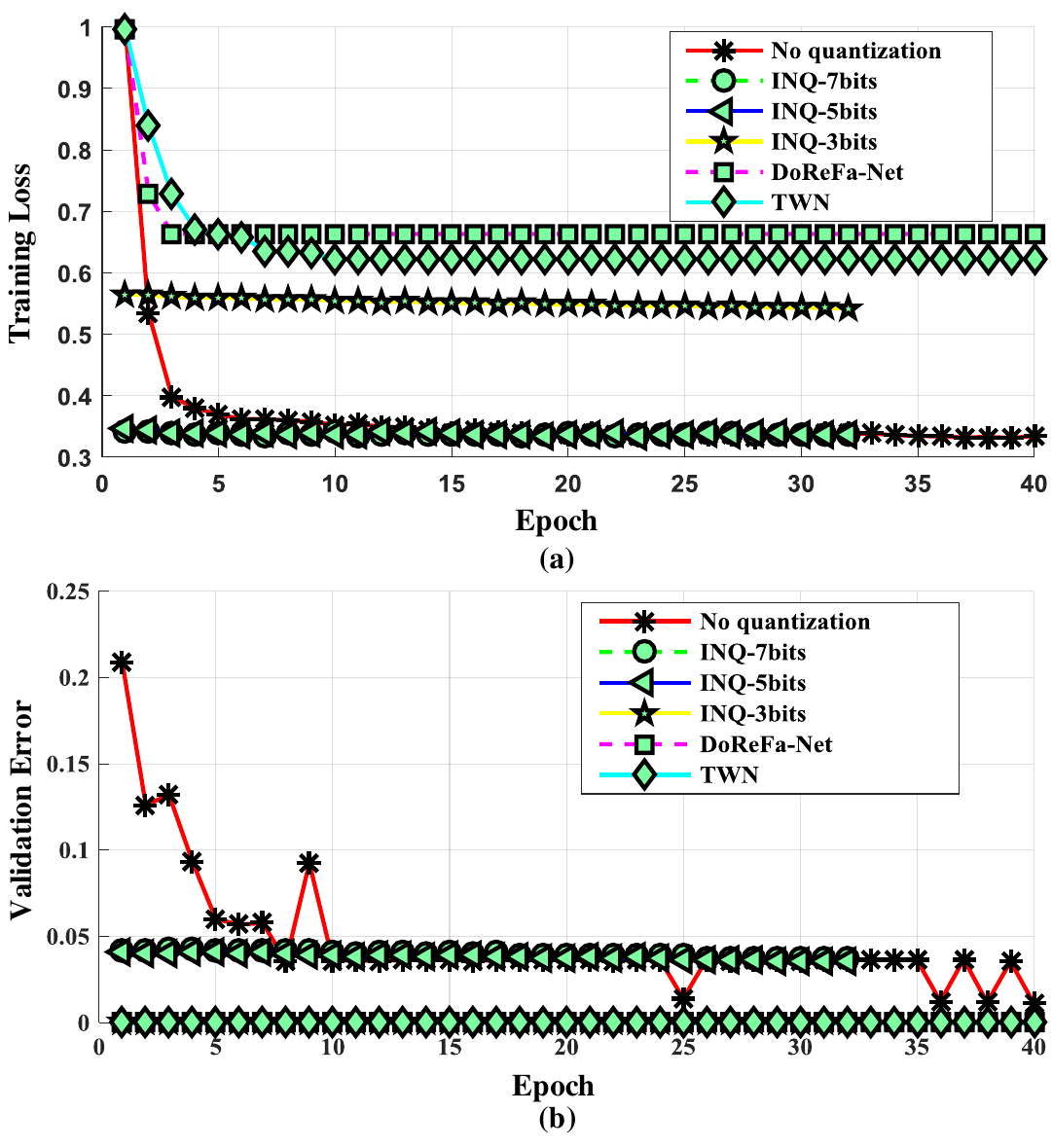}}
\end{center}
\vspace{-20pt}
   \caption{(a) Learning loss and (b) validation error on the Gland dataset with various quantization methods.}
\label{fig:loss}

\end{figure}
\begin{figure}%[ht]
\begin{center}
\centerline{\includegraphics[width=0.6\columnwidth]{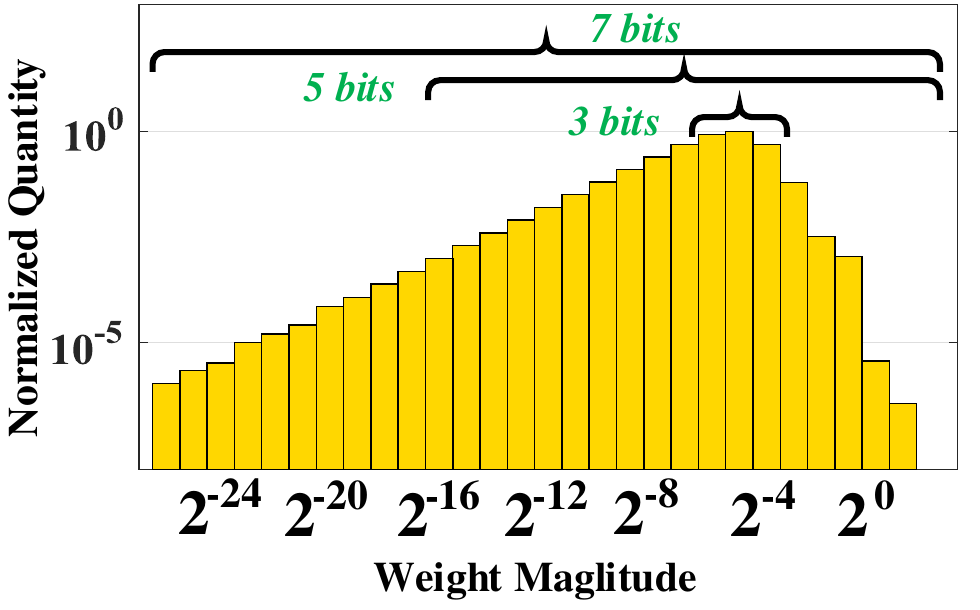}}
\end{center}
\vspace{-20pt}
   \caption{Magnitude distribution of weights in FCNs.}
\label{fig:distribution}
\end{figure}

\vspace{-1pt}
\section{Experiment and Results}
\subsection{Experiment Setup}
We adopt the 2015 MICCAI Gland Challenge dataset \cite{sirinukunwattana2017gland} which have 85 training images ( Part A: 37 normal glands, and Part B: 48 abnormal glands) and 80 testing images (Part A: 60 normal glands, and Part B: 20 abnormal glands).
In suggestive annotation, 16 images with the highest uncertainty scores are extracted first, and then 8 images are collected based on their representativeness using similarity, which are added to the suggested training set in each iteration.
Totally there are 120 iterations in suggestive annotation, and totally 960 suggested training samples are produced.
5 FCNs are used in suggestive annotation, and the waiting time between two annotation suggestion stages is about 10 minutes on a workstation with 4 NVIDIA Tesla P100 GPUs.
We adopt a simple learning rate scaling strategy: set learning rate to 0.0005 in the initial stage, and to 0.00005 when the iteration times reaches a threshold.
As the training time is long, all the configurations are repeated 4 times and the best ones are selected for comparison.
%For model ensemble, the average of outputs from five FCNs are used for evaluation in work \cite{yang2017suggestive}.
%However, in this work, parameter quantity is a big concern and only one FCN is used for quantization.
%We also discuss our method on five FCNs for generality.

%Note that the suggested training samples with incremental quantization have already been discussed in Section \ref{samples}, and in this section we focus on the accuracy of quantized neural networks using suggested training samples.

We will discuss three aspects in the experiment regarding quantization of suggestive annotation (SA), number of parallel FCNs in suggestive annotation, and quantization of network training (NT).
Note that without explicit specifications, one FCN is used in training for segmentation.
All the experiments are evaluated considering detection (F1 score), segmentation (dice score) and shape similarity (object Hausdorff distance) \cite{sirinukunwattana2017gland}.
Several widely-used quantization methods are discussed: incremental quantization, DoReFa-Net, and TWN.
We first perform a simple FCN training with the above quantization methods.
For incremental quantization, we first analyzed the distribution of the weights as shown in Figure \ref{fig:distribution}, and select three configurations: 7 bits, 5 bits, and 3 bits.
As shown in Figure \ref{fig:loss}, it can be noticed that only incremental quantizations with 7 bits and 5 bits have low training loss and achieve comparable performance on the validation dataset with unquantized networks.
Incremental quantization with 3 bits, DoReFa-Net, and TWN obtain a large training loss, and their validation accuracy is almost zero.
Though this is common in network quantization, the accuracy degradation of FCNs is much larger compared with general DNNs, which is possibly due to the following two reasons.
First, unlike general DNNs, FCNs has no fully connected layers resulting in less redundance.
Second, the performance of segmentation is determined in the object level, which means successful segmentation requires correct classification of a doze of pixels in an object.
This is much harder than general classifications using DNNs.
Considering the above discussions, we adopt incremental quantization with 7 bits and 5 bits in the rest of the experiments.

%Details setup for each quantization methods are discussed in the corresponding subsections.

\begin{figure*}[!htbp]
\begin{center}
\centerline{\includegraphics[width=2\columnwidth]{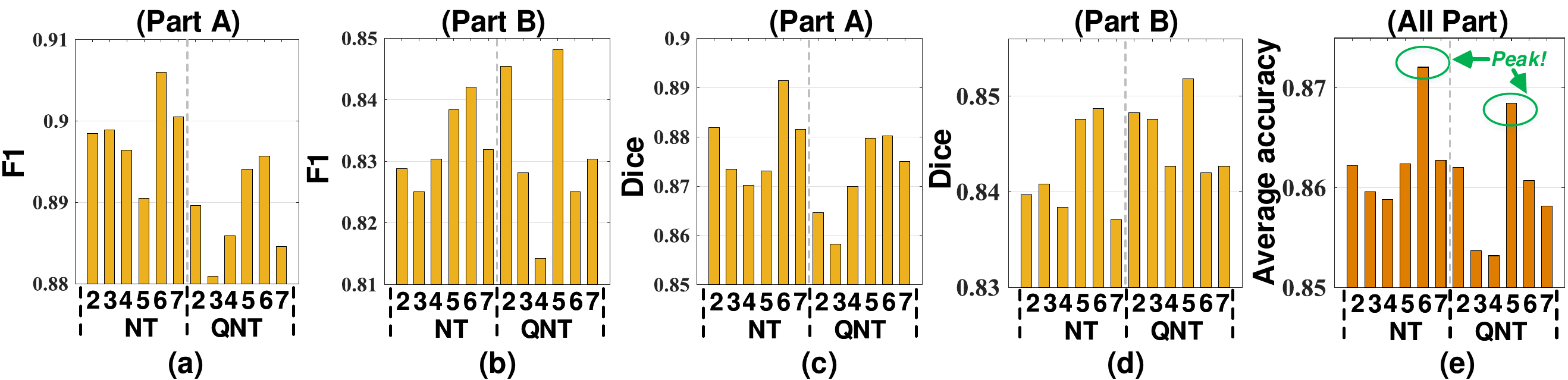}}
\end{center}
   \vspace{-25pt}
   \caption{Performance comparison of \textbf{different number (2,3,4,5,6,7) of suggestive FCNs} with \textbf{network training (NT)} and \textbf{network training with quantization (QNT)}. The QNT is quantized using INQ with 7 bits.
   %The green dash-line corresponds to the performance of the work \cite{yang2017suggestive} with the same configuration.
   }
\label{fig:parallel}
\end{figure*}

\begin{figure*}[!htbp]
\begin{center}
\centerline{\includegraphics[width=2\columnwidth]{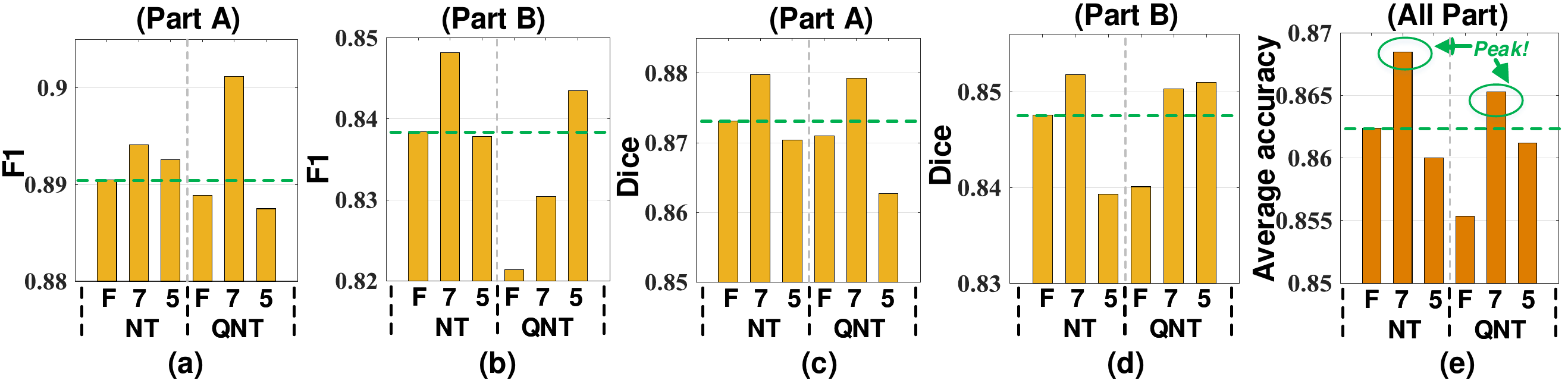}}
\end{center}
\vspace{-25pt}
   \caption{Performance comparison of quantization methods (\textbf{F: float, 7: INQ with 7 bits, 5: INQ with 5 bits}) in \textbf{suggestive annotation with quantization} with \textbf{network training (NT)} and \textbf{network training with quantization (QNT)}. The QNT is quantized using INQ with 7 bits. The green dash-line corresponds to the performance of the work \cite{yang2017suggestive} with the same configuration.}
\label{fig:quantizationSegmentation}
\end{figure*}

\begin{figure*}[!htbp]
\begin{center}
\centerline{\includegraphics[width=2\columnwidth]{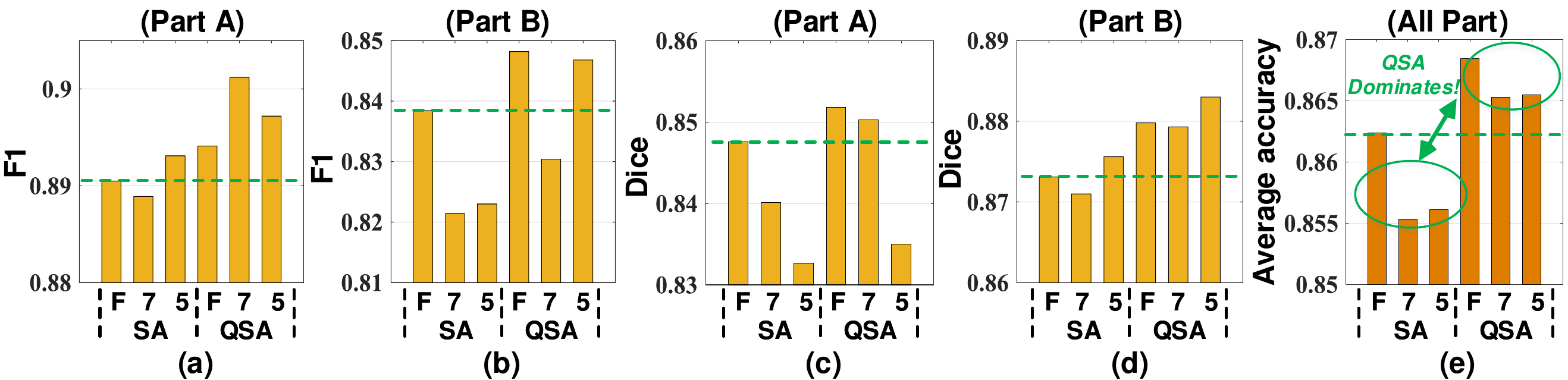}}
\end{center}
\vspace{-5pt}
   \caption{Performance comparison of quantization methods (\textbf{F: float, 7: INQ with 7 bits, 5: INQ with 5 bits}) in \textbf{network training with quantization} with training samples from suggestive annotation (SA) and suggestive annotation with quantization (QSA). The QSA is quantized using INQ with 7 bits. The green dash-line corresponds to the performance of the work \cite{yang2017suggestive} with the same configuration.}
\label{fig:quantizationSuggestive}
%\vspace{-5pt}
\end{figure*}

\begin{figure*}[!htbp]
\begin{center}
\centerline{\includegraphics[width=2\columnwidth]{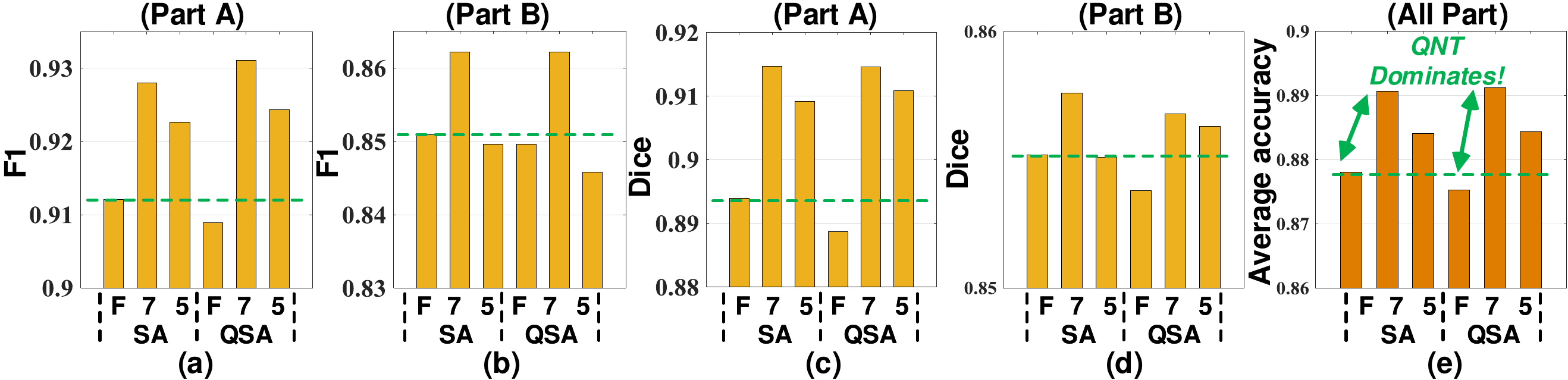}}
\end{center}
\vspace{-5pt}
   \caption{Performance comparison of quantization methods (\textbf{F: float, 7: INQ with 7 bits, 5: INQ with 5 bits}) in \textbf{network training with quantization} using training samples from suggestive annotation (SA) and suggestive annotation with quantization (QSA). The QSA is quantized using INQ with 7 bits. Note that \textbf{ensemble technique with 5 FCNs} are adopted here. The green dash-line corresponds to the performance of the work \cite{yang2017suggestive} with the same configuration.}
\label{fig:quantizationSegmentation5}
%\vspace{-5pt}
\end{figure*}
%evaluation method:
%Accuracy with FCN-5 and FCN-1, training and validation wave, and weight distribution are presented;

%We use 5\% of training data as validation set to select the best model

%Dataset: Gland and lymph node ultrasound image
%Include results from other papers?
\begin{table*}[!htbp]
\centering
%\vspace{18pt}
\caption{Performance comparison with existing works using \textbf{five} FCNs on the MICCAI Gland dataset. The work \cite{yang2017suggestive} achieves state-of-the-art performance on the dataset.}
%\vspace{18pt}
\label{my-label}
\begin{tabular}{ccccccc}
\hline
\multirow{2}{*}{Configuration} & \multicolumn{2}{c}{F1 Score} & \multicolumn{2}{c}{Object Dice} & \multicolumn{2}{c}{Object Hausdorff} \\ \cline{2-7}
                                           & Part A  & Part B  & Part A   & Part B    & Part A    & Part B   \\ \hline
SA (5 FCNs + INQ-7bits) + NT (5 FCNs + INQ-7bits)
                 &   \textbf{0.930}	&\textbf{0.862}&	\textbf{0.914}	&\textbf{0.859}     & \textbf{41.783}     & 97.390        \\
Suggestive annotation \cite{yang2017suggestive}    &  0.921&	0.855&	0.904&	0.858     & 44.736     & \textbf{96.976}                  \\
%Suggestive annotation \cite{yang2017suggestive} (run in our setup)
%               & 0.912	&{0.851}&	{0.894}&	{0.855}     & 44.596      & 98.957            \\
Multichannel \cite{xu2017gland}   &  0.893&	0.843&	0.908	&0.833    & 44.129       & 116.821            \\
Multichannel \cite{xu2016gland}   &  0.858&	0.771&	0.888	&0.815    & 54.202       & 129.930            \\
CUMedVision  \cite{chen2016dcan}   &  0.912&	0.716&	0.897	&0.781    & 45.418       & 160.347            \\  \hline
\end{tabular}
\label{tab:inq-f5}
\end{table*}
\subsection{Impact of Number of Parallel FCNs}

We first discuss the impact of number of parallel FCNs in suggestive annotation.
As shown in Figure \ref{fig:parallel}, six configurations are discussed.
We find that the same trend exists in all configurations: a moderate accuracy is obtained with number of 2, and then the accuracy decreases, and a local minimum occurs with the number of around 4; then the accuracy will increase to a local maximum and decrease afterwards.
It seems that there exists much redundance in FCNs for suggestive annotation, and proper number of parallel FCNs will contribute to the performance.
We will adopts 5 parallel FCNs in suggestive annotation in the experiments afterwards.
In Figure \ref{fig:parallel}(e), we can find that network training can achieve higher accuracy in most of the configurations compared with network training with quantization.
That is to say quantization of network training will hurt the accuracy for some configurations.

\vspace{6pt}
\subsection{Discussion on Suggestive Annotation Quantization}
%We adopt three weight-bit configurations for incremental quantization: 7 bits, 5 bits and 3 bits.
%In order to determine the range of weights, we quantize all the weights with floating point representation and its distribution is shown in Figure \ref{fig:distribution}.
%Note that the figure is in log scale and most of the weights is in the range between $2^{-5}$ and $2^{-3}$.

As shown in Figure \ref{fig:quantizationSegmentation}, suggestive annotation with INQ with 7 bits can always obtain higher accuracy compared with that with the other two.
This reversed U-shape trend indicates that suggestive annotation with INQ with 7 bits may be close to the best fitting point, and loose quantization (no quantization or floating-point representation) and tight quantization (INQ with 5 bits) both degrade the fitting and accuracy loss arises.
By comparing network training and network training with quantization, we can find that network training with quantization will not always improve the accuracy.

\vspace{6pt}
\subsection{Discussion on Network Training}

As shown in Figure \ref{fig:quantizationSuggestive}, unlike suggestive annotation with quantization, the highest accuracy of network training with quantization is achieved with floating-point representation in most of the configurations.
This means network training with quantization will degrade the performance.
By Comparing network training and network training with quantization, we can notice that suggestive annotation with quantization has a great contribution to performance improvement, and the average improvement is 0.9\%.

\vspace{6pt}
\subsection{Comparison with Existing Works}

In order to make fair comparison with existing works, we adopts ensemble methods and set the number of FCNs in network training to 5, which is the same as \cite{yang2017suggestive}.
Several configurations are evaluated as shown in Figure \ref{fig:quantizationSegmentation5}.
Suggestive annotation with quantization shows the same trend as network training with quantization.
In Figure \ref{fig:quantizationSegmentation} and Figure \ref{fig:quantizationSuggestive}, suggestive annotation with quantization has a great impact on the performance with one FCN, while network training with quantization has a significant influence on the performance with five FCNs.
This is due to the fact that the network behaviour of multiple networks with ensemble methods differs from that of only one network.

Comparison with existing works are shown in Table \ref{tab:inq-f5}.
With proper quantization techniques, our proposed method can achieve the best performance on all aspects except object Hausdorff distance on part B.
%As limited repeat times (4 in our experiment), we can only get a suboptimal result of the work
For part A with nonmalignant subjects, out methods can achieve a 0.9\%-1\% improvement with the current state-of-the-art method.
For part B with malignant subjects, it is much harder to segment, and our method gets a 0.1\%-0.7\% improvement.
%The average improvement is 0.68\%.
We achieve comparable performance on object Hausdorff distance on part B, which is only 0.4\% worse than suggestion annotation.
In addition, our method can also obtain 4,6x and 6.4x reduction on memory usage for INQ with 7 bits and 5 bits, respectively.
As activations are in floating point representation, the runtime are not affected.

%1 table
%\section{Discussion}
%generalize to other networks?

\section{Conclusion}
Usually quantization is used to reduce the bit length of parameters with some accuracy loss.
In this paper, we apply quantization to FCNs for accurate biomedical image segmentation, and quantization is used to reduce overfitting in FCNs.
Particularly we base our work on current the state-of-the-art work \cite{yang2017suggestive}, and it has two steps: suggestive annotation and network training.
We add two quantization processes to the two steps, respectively: one to suggestive annotation for high-representative training samples, and the other to general training for high accuracy.
%Suggestive annotation is a critical step which extracts representative samples from the raw training samples.
%In this paper, we first apply general quantization to FCNs to compress model.
%Note that unlike general DNNs, FCNs have no full connection layers.
%Second, we go a step further and add a quantization processing to the current suggestive annotation, which yields a better samples for training.
%Results with a variety of quantization methods show that our approach can achieves much better accuracy than general quantized FCNs.
Extensive experiments are presented on the widely-used MICCIA Gland dataset.
Results show that both quantization processes can improve the segmentation performance by around 1\% for some configurations.
However, for specific networks, usually there is only one process dominates in the performance.
For network training with only one FCN, suggestive annotation with quantization dominates, while network training with quantization dominates for network training with five FCNs.
The number of parallel FCNs in suggestive annotation will also affect the performance.
Our proposed method exceeds the current state-of-the-art performance by up to 1\%.
In addition, our method has a up to 6.4x reduction on memory usage.

{\small
\bibliographystyle{ieee}

}

\end{document}